\documentclass[sigconf]{acmart}
\AtBeginDocument{%
  }

\usepackage{booktabs}
\usepackage{cleveref}
\usepackage{graphicx}
\usepackage{multicol}
\usepackage{multirow}
\usepackage{colortbl}
\usepackage{amsmath}
\usepackage[ruled,linesnumbered]{algorithm2e}
\usepackage{enumitem}
\usepackage{subfig}

\newcommand{\R}[1]{\textcolor{red}{#1}}
\newcommand{\B}[1]{\textcolor{blue}{#1}}
\newcommand{\DF}[1]{\textcolor{gray}{#1}}

\def \submission {}

\usepackage{xspace}
\usepackage{comment}
\usepackage{lipsum}

\makeatletter
\DeclareRobustCommand\onedot{\futurelet\@let@token\@onedot}
\def\@onedot{\ifx\@let@token.\else.\null\fi\xspace}

\def\ie{\emph{i.e}\onedot}

\usepackage{listings}
\usepackage{float}

\definecolor{codegreen}{rgb}{0,0.6,0}
\definecolor{codegray}{rgb}{0.5,0.5,0.5}
\definecolor{codepurple}{rgb}{0.58,0,0.82}
\definecolor{backcolour}{rgb}{0.95,0.95,0.92}

\lstdefinestyle{mystyle}{
    backgroundcolor=\color{backcolour},   
    commentstyle=\color{codegreen},
    keywordstyle=\color{magenta},
    numberstyle=\tiny\color{codegray},
    stringstyle=\color{codepurple},
    basicstyle=\ttfamily\footnotesize,
    breakatwhitespace=false,         
    breaklines=true,                 
    captionpos=b,                    
    keepspaces=true,                 
    numbers=left,                    
    numbersep=5pt,                  
    showspaces=false,                
    showstringspaces=false,
    showtabs=false,                  
    tabsize=2
}

\newfloat{lstfloat}{htbp}{lop}
\floatname{lstfloat}{Algorithm}

\lstdefinestyle{myverbatim}{
    basicstyle=\ttfamily\footnotesize,
    backgroundcolor=\color{white},
    breaklines=true,
    breakatwhitespace=true
}

\definecolor{Crimson}{rgb}{0.86, 0.08, 0.24}
\definecolor{DarkGreen}{rgb}{0.00, 0.40, 0.00}
\definecolor{RoyalBlue}{rgb}{0.25, 0.35, 0.74}
\definecolor{DarkCyan}{rgb}{0.0, 0.54, 0.54}
\definecolor{Gray}{gray}{0.95}
\definecolor{ChromeYellow}{rgb}{1.0, 0.65, 0.0}
\definecolor{Burgundy}{RGB}{167, 121, 121}
\definecolor{Red}{rgb}{1.0, 0.0, 0.0}

\ifx \submission \undefined
\newcommand{\wu}[1]{\textcolor{DarkGreen}{#1}}

\newcommand{\lin}[1]{\textcolor{DarkCyan}{#1}}

\else
\newcommand{\wu}[1]{{#1}}

\newcommand{\lin}[1]{{#1}}

\fi

\setcopyright{acmlicensed}
\copyrightyear{2025}
\acmYear{2025}
\setcopyright{rightsretained}
\acmConference[MM '25] {Proceedings of the 33rd ACM International Conference on Multimedia}{October 27--31, 2025}{Dublin, Ireland.}
\acmBooktitle{Proceedings of the 33rd ACM International Conference on Multimedia (MM '25), October 27--31, 2025, Dublin, Ireland}
\acmDOI{10.1145/3746027.3754939}
\acmISBN{979-8-4007-2035-2/2025/10}

\begin{document}

\title{InstructFLIP: Exploring Unified Vision-Language Model for Face Anti-spoofing}


\author{Kun-Hsiang Lin}
\orcid{0009-0009-8040-979X}
\affiliation{%
  \institution{National Taiwan University}
  \city{Taipei}
  \country{Taiwan}
}
\email{jacklin@cmlab.csie.ntu.edu.tw}

\author{Yu-Wen Tseng}
\orcid{0009-0001-8332-8444}
\affiliation{%
  \institution{National Taiwan University}
  \city{Taipei}
  \country{Taiwan}
}
\email{d12922018@csie.ntu.edu.tw}

\author{Kang-Yang Huang}
\orcid{0000-0003-1268-5214}
\affiliation{%
  \institution{National Taiwan University}
  \city{Taipei}
  \country{Taiwan}
}
\email{huangkangyang@cmlab.csie.ntu.edu.tw}

\author{Jhih-Ciang Wu}
\orcid{0000-0003-4071-3980}
\affiliation{%
  \institution{National Taiwan Normal University}
  \city{Taipei}
  \country{Taiwan}
}
\email{jcwu@csie.ntnu.edu.tw}

\author{Wen-Huang Cheng}
\orcid{0000-0002-4662-7875}
\authornote{Corresponding author}
\affiliation{%
  \institution{National Taiwan University}
  \city{Taipei}
  \country{Taiwan}
}
\email{wenhuang@csie.ntu.edu.tw}

\renewcommand{\shortauthors}{Lin et al.}

\begin{abstract}
  Face anti-spoofing (FAS) aims to construct a robust system that can withstand diverse attacks. While recent efforts have concentrated mainly on cross-domain generalization, two significant challenges persist: {\em limited semantic understanding of attack types} and {\em training redundancy across domains.} We address the first by integrating vision-language models (VLMs) to enhance the perception of visual input. For the second challenge, we employ a meta-domain strategy to learn a unified model that generalizes well across multiple domains. Our proposed {\em InstructFLIP} is a novel instruction-tuned framework that leverages VLMs to enhance generalization via textual guidance trained solely on a single domain. At its core, InstructFLIP explicitly decouples instructions into content and style components, where content-based instructions focus on the essential semantics of spoofing, and style-based instructions consider variations related to the environment and camera characteristics. Extensive experiments demonstrate the effectiveness of InstructFLIP by outperforming SOTA models in accuracy and substantially reducing training redundancy across diverse domains in FAS. Project website is available at~\url{https://kunkunlin1221.github.io/InstructFLIP}.
\end{abstract}

\ccsdesc[500]{Computing methodologies~Biometrics}

\keywords{Face Anti-spoofing, Vision-language models, Unified model}


\maketitle

\section{Introduction}
\label{sec:intro}

Modern face recognition systems have become pervasive, from unlocking smartphones to accessing secure facilities, due to advancements in deep learning and their non-contact, user-friendly nature. However, these systems are susceptible to presentation attacks, where adversaries attempt to bypass security mechanisms using tools such as printed photos, replayed videos, and other advanced deception techniques. Consequently, FAS plays a crucial role in safeguarding these systems against such vulnerabilities. While existing FAS methods exhibit strong performance in intra-dataset evaluations, where training and testing data share the same domain, their effectiveness significantly declines in cross-dataset scenarios due to domain distribution shifts~\cite{yu2022deep}.

\begin{figure}[t]
    \includegraphics[width=0.47\textwidth]{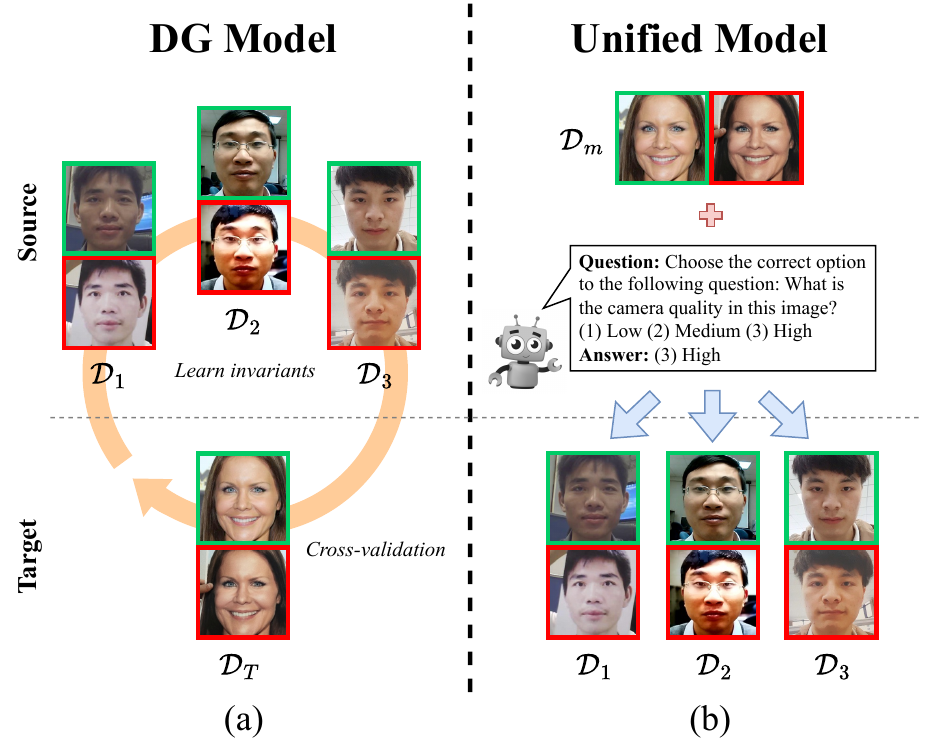}
    \centering
    \footnotesize
    \caption{\wu{Comparison between DG-based and unified model. (a) Conventional DG-based approaches using leave-one-out protocols from multiple domains ($\mathcal{D}_1$, $\mathcal{D}_2$, $\mathcal{D}_3$) to the target one ($\mathcal{D}_T$) result in significant training redundancy by treating each domain independently during cross-validation. (b) We propose InstructFLIP, a unified model that leverages the meta domain $\mathcal{D}_m$ and instruction-based training to jointly learn domain-invariant content and style features, allowing efficient generalizations without redundant retraining.}}
    \Description{The comparison highlights the differences between conventional domain generalization(DG) based face anti-spoofing (FAS) methods and the proposed InstructFLIP approach. (a) Traditional DG-based methods rely on leave-one-out protocols, where multiple source domains ($\mathcal{D}_1, \mathcal{D}_2, \mathcal{D}_3$) are used independently to train models for the target domain ($\mathcal{D}_T$). This process results in significant training redundancy, as each domain is treated separately during cross-validation. (b) In contrast, InstructFLIP introduces a unified model that leverages a meta domain ($\mathcal{D}_m$) and instruction-based training. This approach enables the model to simultaneously learn domain-invariant content and style features, facilitating efficient generalization to new domains without the need for redundant retraining.}
    \label{fig:teaser}
\end{figure}

To ensure the reliability of FAS in real-world applications, considerable research has concentrated on domain generalization (DG) techniques~\cite{wang2020cross, jia2020single, dong2021open, wang2022domain, du2022energy, zhou2022adaptive, sun2023rethinking}. DG aims to facilitate models in learning generalized feature representations that effectively distinguish between genuine and spoofed inputs across diverse and unpredictable environments, thereby enhancing both robustness and adaptability. Such approaches are indispensable for developing systems capable of maintaining high performance even in the presence of novel, previously unseen attack scenarios. However, DG requires extensive cross-validation across datasets, resulting in training redundancy due to repeated retraining. In Figure~\ref{fig:teaser}(a), conventional DG methods exacerbate this issue by using leave-one-out protocols, where models are trained on multiple domains like ($\mathcal{D}_1$, $\mathcal{D}_2$, $\mathcal{D}_3$) and evaluated on a target domain ($\mathcal{D}_T$), treating each domain independently. This inefficiency limits the practicality of DG for large-scale deployment as they struggle to produce a unified, generalizable model. Additionally, DG often lacks transparency, providing limited insight into their decision-making processes.

Recent advancements in VLMs, such as CLIP~\cite{radford2021learning}, aim to improve generalization through cross-modal learning. For example, FLIP~\cite{srivatsan2023flip} leverages the pretrained capabilities of CLIP to address the FAS task, while CFPL~\cite{liu2024cfpl} employs BLIP2's Q-Former~\cite{li2023blip} to bridge the modality gap between images and pretrained text encoders. However, these VLM-based methods have not fully exploited the rich semantic information inherent in language, limiting their effectiveness in complex scenarios requiring meticulous understanding.

In summary, two fundamental limitations remain unaddressed in existing DG and VLM approaches: (1) \textit{limited semantic understanding of facial attacks}, as many methods struggle to accurately understand facial attacks due to interference from factors such as environmental conditions and camera quality. While most rely on implicit patterns to propose solutions, they lack intuition, making reliable decision-making challenging; and (2) \textit{training redundancy across domains}, where traditional leave-one-out protocols require models to be trained and validated independently on multiple domains, leading to excessive retraining. To address these challenges, we propose InstructFLIP, a novel VLM-based architecture that applies instruction tuning to explicitly extract and encode high-level semantic priors from textual descriptions, as illustrated in Figure~\ref{fig:teaser}(b). Effective FAS necessitates the learning of domain-invariant representations while preserving sensitivity to discriminative, non-spoof-specific attributes. By conditioning the model on structured linguistic instructions, InstructFLIP enhances its capacity to align visual features with task-relevant semantics, thereby improving generalization and robustness across heterogeneous domains.

To more effectively encode spoof-related and non-spoof-related attributes within the VLM framework, we decompose linguistic supervision into two orthogonal components: content and style. The content component encodes spoof-specific semantics such as replay, print, mask attacks, etc., while the style component captures nuisance factors unrelated to spoofing, including image quality, environmental context, and illumination conditions. This structured decomposition facilitates disentangled feature learning and enhances the model’s robustness to domain shifts. Aiming to avoid training redundancy, we introduce a richly labeled meta-domain $\mathcal{D}_m$ designed to synthesize diverse image-instruction pairs for efficient and unified model training. As depicted in Figure~\ref{fig:teaser}(b), our approach leverages $\mathcal{D}_m$ in conjunction with instruction tuning to jointly learn domain-invariant content and style representations, eliminating the need for repeated retraining across domains. This paradigm significantly reduces training overhead while improving generalization capability, thereby enhancing the model’s scalability for real-world deployment. By consolidating diverse supervisory signals into a single training cycle, InstructFLIP effectively captures multiple task-relevant factors, enabling adaptive and generalized FAS. The main contributions of this paper are as follows:
\begin{itemize}[leftmargin=*]
    \item This paper proposes InstructFLIP, a novel instruction-tuned VLM framework for FAS, which integrates textual supervision to enhance semantic understanding of spoofing cues.
    \item We design a content-style decoupling mechanism that explicitly separates spoof-related (content) and spoof-irrelevant (style) information, improving generalization to unseen domains.
    \item We introduce a meta-domain learning strategy to eliminate training redundancy in cross-domain settings by utilizing diverse image-instruction pairs sampled from a structured meta-domain.
    \item Experimental results demonstrate that InstructFLIP surpasses SOTA methods across multiple FAS benchmarks, effectively capturing spoof-related patterns through language-guided supervision while substantially reducing training overhead, thereby enhancing its applicability in real-world scenarios.
\end{itemize}

\section{Related Work}
\label{sec:relate}
\subsection{Face Anti-Spoofing} 
FAS is a vital safeguard in face recognition systems aiming to mitigate the risks posed by presentation attacks. Over recent decades, research in this field has significantly evolved, particularly with the advancement of deep learning. The focus has shifted towards leveraging neural networks to enhance detection accuracy. This includes the development of classification-based models~\cite{yang2014learn,patel2016secure,amin2018face,liu2018learning,yang2019face} and other auxiliary-learning methods~\cite{yu2020cvpr, wang2022patchnet}, and generative frameworks~\cite{stehouwer2020noise, wang2022disentangled} to improve robustness against spoofing attempts.

The increasing complexity of cross-domain variability, driven by diverse application scenarios and heterogeneous datasets, has led to the development of domain adaptation (DA) techniques~\cite{zhou2022generative, yue2023cyclically, wang2020unsupervised}. These methods enhance model adaptability by learning shared feature representations across source and target domains. However, DA approaches often require extensive fine-tuning and substantial labeled data, which limits their scalability and practicality for real-world FAS applications. To overcome these challenges, DG techniques~\cite{zhang2020casia, wang2020cross, jia2020single, wang2022domain, zhou2023instance, sun2023rethinking} have been developed to extract domain-invariant features, enabling models to generalize to unseen domains. Despite their potential, DG methods necessitate extensive cross-validation across multiple datasets, resulting in significant training inefficiency and high costs due to repeated retraining. This constraint hampers their practicality for large-scale deployment, as they not only fail to produce a unified, adaptable model but also often lack the reasoning ability needed for reliable decision-making.

Therefore, we propose InstructFLIP, which eliminates the need for redundant retraining by utilizing a single meta-domain for unified model learning. By incorporating instruction tuning with rich linguistic guidance, InstructFLIP achieves improved semantic alignment and robust generalization across diverse domains.



\subsection{Visual Language Models} 
Since the introduction of CLIP~\cite{radford2021learning}, VLMs~\cite{li2022blip, dai2023instructblip} have advanced rapidly, yielding significant breakthroughs across multiple research fields~\cite{zhang2022tip, zhou2022conditional, gu2022openvocabulary}. In the context of FAS, researchers have explored the integration of textual information to facilitate DG. For instance, FLIP~\cite{srivatsan2023flip} pioneered the application of CLIP to address FAS challenges, while extensions such as ~\cite{liu2024cfpl, liu2024bottom, hu2024fine, liu2024fmclip, zou2024lasoftmoe, guo2024style, chen2025moae, zhang2025interpretable} sought to enhance the generalization capabilities of FAS models through visual language learning. Nevertheless, these approaches have primarily targeted cross-domain issues without fully leveraging the potential of textual understanding to improve model understanding and performance in unseen scenarios.

To address the limitations of existing approaches, this work presents InstructFLIP, a novel instruction-tuned methodology inspired by InstructBLIP~\cite{dai2023instructblip}. This approach leverages the rich annotations available in the CelebA-Spoof (CS) dataset ~\cite{2020ZhangCelebA-Spoof} to construct structured instruction-based image-text pairs, enabling the training of a unified FAS model. By incorporating comprehensive and semantically meaningful instructions, the model learns to generalize effectively across diverse domains while gaining language-level representations of spoofing mechanisms. This not only enhances robustness and improves the accurate identification of spoof types, but also underscores the critical role of reasoning capabilities in VLM for effective FAS solutions.


\begin{figure*}
    \centering
    \includegraphics[width=0.93\textwidth]{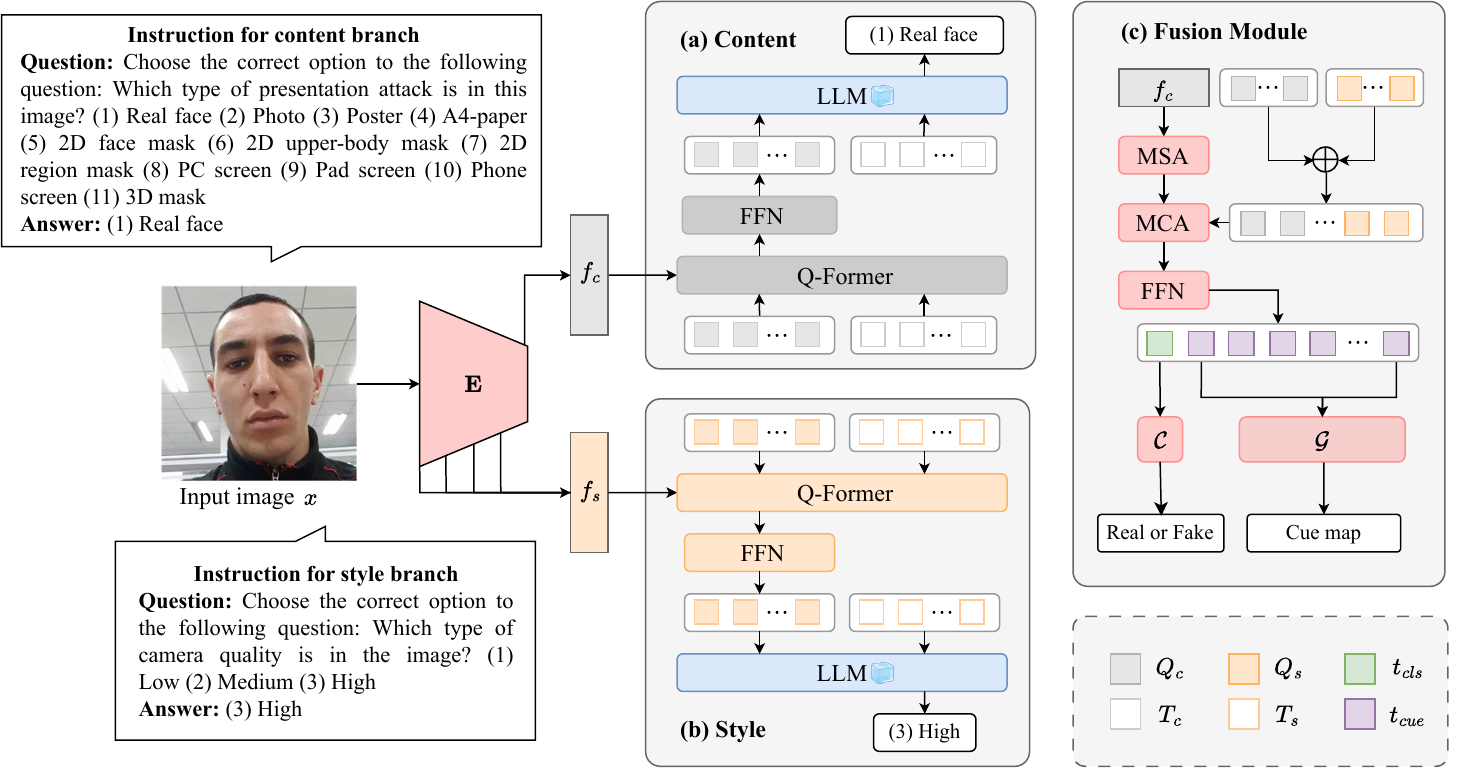}
    \caption{Overview of the proposed InstructFLIP framework for FAS. We first encode the input image $x$, resulting in the separated content and style features $f_c$ and $f_s$. The dual-branch architecture presented in (a) and (b) is designed for instruction tuning based on the corresponding expertise in each branch. Eventually, the prediction for determining $x$ is spoofing or not is carried out in (c) via a classifier $\mathcal{C}$ according to the fused queries and $f_c$, coupling with the cue map generated by $\mathcal{G}$.}
    \Description{The proposed InstructFLIP framework for face anti-spoofing (FAS) is designed to effectively separate and utilize content and style features for robust spoof detection. First, the input image $x$ is encoded to extract content features ($f_c$) and style features ($f_s$). The framework employs a dual-branch architecture, as shown in (a) and (b), where each branch is specialized for instruction tuning based on its respective expertise. Finally, the prediction to determine whether $x$ is spoofed or genuine is performed in (c) using a classifier ($\mathcal{C}$). This decision is based on the fused queries and $f_c$, along with the cue map generated by the generator ($\mathcal{G}$), ensuring accurate and interpretable spoof detection.}
    \label{fig:framework}
\end{figure*}

\section{Proposed Method}
\label{sec:method}

\begin{table}[t]
\centering
\caption{Options of content-related and style-related prompts.}
\label{tab:prompts}
\vspace{-4pt}
\scalebox{0.91}{
    \begin{tabular}{|c|c|p{4.8cm}|}
    \hline
    \rowcolor{lightgray} \textbf{Type} & \textbf{Question} & \multicolumn{1}{c|}{\textbf{Options}} \\
    \hline
    \hline
    \multirow{5}{*}{Content} & 
    \multirow{5}{*}{Spoof type} & 
    (1) Real face (2) Photo (3) Poster (4) A4-paper (5) 2D face mask (6) 2D upper-body mask (7) 2D region mask (8) PC screen (9) Pad screen (10) Phone screen (11) 3D mask \\
    \hline
    \multirow{3}{*}{Style} & Illumination  & (1) Normal (2) Strong (3) Back (4) Dark \\
        \cline{2-3}
       & Environment    & (1) Indoor (2) Outdoor \\
        \cline{2-3}
       & Camera quality & (1) Low (2) Medium (3) High \\
    \hline
    \end{tabular}
}
\vspace{-8pt}
\end{table}

Learning to perform FAS often presents DG challenges, requiring significant training redundancy to accommodate diverse protocol requirements. To address this, we aim to develop a unified approach capable of effectively adapting to various scenarios. Specifically, we propose \textbf{Instruct}-tuned \textbf{F}ace Anti-spoofing \textbf{L}anguage-\textbf{I}mage \textbf{P}re-training (InstructFLIP), a novel framework fine-tuned on a meta-domain $\mathcal{D}_m$, which enables DG without the need for repeated retraining, as illustrated in Figure~\ref{fig:framework}. By decoupling the representation into content and style features, our model can adapt to varying domain characteristics by leveraging corresponding interpretable components, promoting a comprehensive visual understanding that enhances generalization to handle domain variations. The learned embeddings from the content and style branches are further fused with the visual encoder via a dedicated fusion module to produce the final prediction, enabling LLM-free inference and supporting real-time deployment. In the following section, we elaborate on our InstructFLIP in addressing FAS, which facilitates robust generalization across diverse domains.

\subsection{Instructive Prompts}
To efficiently capture versatile semantic representations for dealing with diverse domains, we restrict our training data solely from $\mathcal{D}_m$~\cite{2020ZhangCelebA-Spoof}, enable to learn a unified model that generalizes effectively. By leveraging the instances from $\mathcal{D}_m$, we aim to build a model capable of learning domain-invariant features, which facilitates robust performance across varying domain conditions without needing extensive retraining on each specific domain.

Building on insights from previous literature~\cite{wang2022domain,zhou2023instance,liu2024cfpl}, we structure the instructive prompts into two categories, \ie, content and style prompts. The former targets to identify the spoofing types, while the latter focuses on style conditions, such as illumination, environment, and camera quality\lin{, as introduced by $\mathcal{D}_m$}. More precisely, we craft instruction-based questions for content prompts, such as ``Choose the correct option for the following question: Which spoof type is in this image?'' The options provided cover a range of spoofing types along with the real face category, as outlined in Table~\ref{tab:prompts}. For style prompts, we use a similar structure to content prompts but concentrate on style attributes. For example:  ``Choose the correct option for the following question: What is the illumination condition/environment/camera quality in this image?'' The options here are tailored to address variations in lighting, background, and camera quality, as indicated by style questions. By exploring $\mathcal{D}_m$ with these prompts, we create a varied set for enriching texture inputs that are crucial while training the unified model. The composition of content and style prompts ensures that the model is equipped with a comprehensive understanding of both the visual characteristics that influence spoofing detection, improving its general performance across various environmental conditions.

\subsection{Visual Content and Style Features}
Since the instruction prompts are divided into content and style classes, we extend this concept to disentangle features. Given a face image $x$, we represent $f_c = \mathbf{E}(x)$ as the content feature encoded by a feature extractor $\mathbf{E}$ and obtained from the highest-level layer. To construct the style feature, we adopt an approach inspired by Adaptive Instance Normalization~\cite{huang2017arbitrary} to assemble a multi-level feature. Unlike conventional method~\cite{liu2024cfpl} that aggregates style features from each level through averaging, we propose retaining distinct characteristics from each level. Specifically, we concatenate the mean and standard deviation of the features from multiple levels, which can be formulated as
\begin{equation}
    f_s = s_1 \oplus s_2 \cdots \oplus s_L,
\label{eq:visual_features}
\end{equation}
where $s_i = \mu_i \oplus \sigma_i$ stands for a concatenation of the mean and standard deviation in $i$-level. $\mu_i$ and $\sigma_i$ are obtained by calculating the statistic value of $\mathbf{E}_i(x)$.

We clarify the notations by elaborating on the dimensions of $f_c$ and $f_s$ for a clear understanding. The content feature $f_c \in \mathbb{R}^{N\times{d}}$ is a tensor, where $N$ represents the total number of tokens, including the class and patch tokens, and $d$ is the dimensionality of each token embedding. The style feature $f_s \in \mathbb{R}^{2L\times{d}}$ is a multi-level representation,  where $L$ denotes the total number of layers in $\mathbf{E}$.

\subsection{Instruction Tuning with Visual Features}
We employ a couple of features,  \ie, $f_c$ and $f_s$, by proposing a dual-branch architecture that learns distinct representations for each aspect. The content branch attempts to capture the attributes directly related to attack types, while the style branch gathers contextual information not directly associated with spoofing but is vital for understanding scene variability. As depicted in Figure~\ref{fig:framework}, the content and style branches share a similar architectural structure. Here, we primarily describe the process within the content branch, where the equivalent operations can likewise be applied to the style branch. The instructive content prompt first goes through the standard process, such as tokenization and word embedding, resulting in an instruction representation $T_c \in \mathbb{R}^{j\times {d}}$. This instruction $T_c$ is combined with the learnable query $Q_c \in \mathbb{R}^{k\times {d}}$ to compose the input for the Q-Former~\cite{li2023blip}, where $j$ and $k$ represent the length of instruction and the number of learnable queries, respectively. The Q-Former stacks multiple blocks, with each containing multi-head self-attention (MSA) and multi-head cross-attention (MCA), where we symbolize them by $\phi$ and $\psi$ for simplicity. For each hidden state $h$, the MSA is defined as
\begin{equation}
    \tilde{h} = h + \phi(h),
\label{eq:MSA}
\end{equation}
where $h = Q_c \oplus T_c$ is the initial state for the Q-Former, which is subsequently updated. We incorporate the hidden state with content feature via MCA, which can be formulated as
\begin{equation}
    \hat{h} = h + \psi(h,f_c).
\label{eq:MCA}
\end{equation}
To streamline the expression, we omit layer normalization typically applied to each hidden state and the content feature before the attention mechanism and disregard the two linear layers with the activation function expanded after the MCA.

After passing through the Q-Former, the query effectively learns to capture critical affinities from the content feature. The processed query is subsequently concatenated with the instruction representation, serving as a soft prompt to frozen LLMs to generate the corresponding prediction $p_c$, \ie, spoofing type or real face in the content branch.

The objectives in the two branches seek that the model effectively distinguishes between spoof-related and domain-invariant characteristics. For example, the objective in the content branch is optimized using cross-entropy loss, accurately classifying spoof-related attributes by minimizing the discrepancy between prediction and ground truth, which can be formulated as
\begin{equation}
    \mathcal{L}_{c} = -\sum y_c \log(p_c), 
\label{eq:content_loss}
\end{equation}
where $y_c$ and $p_c$ represent the label and prediction, respectively. The style branch is optimized by $\mathcal{L}_{s}$ similarly to~(\ref{eq:content_loss}).

\begin{table*}[t]
\centering
\caption{Unified FAS benchmark results on MCIO and WCS datasets. The CA dataset is used for training, with evaluation conducted on both MCIO and WCS benchmarks. Subtable (a) presents results on MCIO, (b) reports on WCS, and (c) summarizes average performance across all datasets. The best and second-best results are highlighted in \R{red} and \B{blue}, respectively.}
\label{tab:unified}

\subfloat[\textbf{Results on MCIO datasets}]{
    \label{tab:unified_a}
    \setlength{\tabcolsep}{3pt}
    \scalebox{0.91}{
        \begin{tabular}{llccclccclccclccc}
        \toprule
        \multirow{4}{*}{\textbf{Method}} & \multirow{4}{*}{\textbf{Venue}} &
        \multicolumn{3}{c}{\textbf{M}} & &
        \multicolumn{3}{c}{\textbf{C}} & &
        \multicolumn{3}{c}{\textbf{I}} & &
        \multicolumn{3}{c}{\textbf{O}} \\
        \cmidrule{3-5} \cmidrule{7-9} \cmidrule{11-13} \cmidrule{15-17}
        & &
        \multirow{2}{*}{~HTER$\downarrow$~} & \multirow{2}{*}{~AUC$\uparrow$~} & TPR$\uparrow$@ & &
        \multirow{2}{*}{~HTER$\downarrow$~} & \multirow{2}{*}{~AUC$\uparrow$~} & TPR$\uparrow$@ & &
        \multirow{2}{*}{~HTER$\downarrow$~} & \multirow{2}{*}{~AUC$\uparrow$~} & TPR$\uparrow$@ & &
        \multirow{2}{*}{~HTER$\downarrow$~} & \multirow{2}{*}{~AUC$\uparrow$~} & TPR$\uparrow$@ \\
        & & & & FPR=1\% & & & & FPR=1\% & & & & FPR=1\% & & & & FPR=1\% \\
        \midrule
        SSDG-R~\cite{jia2020single}         & CVPR'20 &    24.92 &     82.42 &     16.19 &&    17.84 &     89.77 &     21.36 &&     18.60 &     90.00 &     47.67 &&     25.68 &     81.19 &        22.74 \\
        ViT~\cite{huang2022adaptive}        & ECCV'22 &    12.69 &     91.42 &     67.93 &&     7.36 &     97.09 &     74.18 &&     20.08 &     87.04 &     64.29 &&     25.48 &     82.18 &        40.58 \\
        SAFAS~\cite{sun2023rethinking}      & CVPR'23 &    31.82 &     74.01 &      3.49 &&    29.26 &     75.17 &      4.20 &&     27.57 &     79.23 &     15.90 &&     29.84 &     76.38 &        13.76 \\
        FLIP-MCL~\cite{srivatsan2023flip}   & ICCV'23 &    13.60 &     91.70 &     59.29 &&     5.25 &     98.99 &     91.11 &&     17.37 &     91.46 &     65.18 &&     17.72 &     89.02 &        40.02 \\
        CFPL~\cite{liu2024cfpl}             & CVPR'24 & \B{8.45} & \B{95.48} & \B{73.31} && \B{2.77} & \B{99.53} & \B{94.36} && \B{10.00} & \B{95.90} & \B{80.13} && \B{17.21} & \B{89.30} & \B{31.58} \\
        InstructFLIP                        & -       & \R{5.52} & \R{98.12} & \R{86.62} && \R{1.47} & \R{99.79} & \R{97.36} &&  \R{9.12} & \R{96.17} & \R{86.48} && \R{14.33} & \R{94.79} & \R{64.77} \\
        \bottomrule
        \end{tabular}
    }
}

\begin{minipage}[t]{0.57\textwidth}
\hspace{-30pt}
\subfloat[\textbf{Results on WCS datasets}]{
    \label{tab:unified_b}
    \setlength{\tabcolsep}{3pt}
    \scalebox{0.87}{
        \begin{tabular}{lccclccclccclccc}
        \toprule
        \multirow{4}{*}{\textbf{Method}} &
        \multicolumn{3}{c}{\textbf{W}} & & 
        \multicolumn{3}{c}{\textbf{C}} & & 
        \multicolumn{3}{c}{\textbf{S}} \\
        \cmidrule{2-4} \cmidrule{6-8} \cmidrule{10-12} &
        \multirow{2}{*}{~HTER$\downarrow$~} & \multirow{2}{*}{~AUC$\uparrow$~} & TPR$\uparrow$@ & & 
        \multirow{2}{*}{~HTER$\downarrow$~} & \multirow{2}{*}{~AUC$\uparrow$~} & TPR$\uparrow$@ & & 
        \multirow{2}{*}{~HTER$\downarrow$~} & \multirow{2}{*}{~AUC$\uparrow$~} & TPR$\uparrow$@ \\
        & & & ~FPR=$1\%$~ & & & & ~FPR=$1\%$~ & & & & ~FPR=$1\%$~ \\ 
        \midrule
        SSDG-R~\cite{jia2020single}        &    41.57 &    62.88 &     2.38 &&    19.98 &    87.87 &     20.38 &&    52.41 &    46.87 &     1.09  \\
        ViT~\cite{huang2022adaptive}       &    28.08 &    78.49 &     6.89 &&    12.50 &    94.52 &     49.71 &&    39.81 &    64.62 &     3.65  \\
        SAFAS~\cite{sun2023rethinking}     &    41.69 &    61.11 &     0.97 &&    18.14 &    89.06 &     23.34 &&    55.51 &    41.29 &     0.76  \\
        FLIP-MCL~\cite{srivatsan2023flip}  &    26.96 &    79.96 &    11.96 &&  \B{7.90}& \B{97.54}&     57.31 &&    42.97 &    59.89 &     0.47  \\    
        CFPL~\cite{liu2024cfpl}            & \B{26.85}& \B{80.97}& \B{17.42}&&  \R{5.50}& \R{98.74}&  \R{78.72}&& \B{42.29}& \B{60.60}&  \B{2.37} \\
        InstructFLIP                       & \R{19.51}& \R{89.90}& \R{39.11}&&     8.49 &    96.81 &  \B{70.17}&& \R{30.33}& \R{80.16}& \R{12.07} \\
        \bottomrule
        \end{tabular}
    }
}
\end{minipage}
\begin{minipage}[t]{0.33\textwidth}
\hspace{32pt}
\subfloat[\textbf{Mean metrics computed over all datasets.}]{
    \label{tab:unified_c}
    \setlength{\tabcolsep}{3pt}
    \scalebox{0.87}{
        \begin{tabular}{lccc}
        \toprule
        \multirow{4}{*}{\textbf{Method}} & \multicolumn{3}{c}{\textbf{Avg.}} \\
        \cmidrule{2-4} &
        \multirow{2}{*}{~HTER$\downarrow$~} & \multirow{2}{*}{~AUC$\uparrow$~} & TPR$\uparrow$@ \\
        & & & ~FPR=$1\%$~ \\ 
        \midrule
        SSDG-R~\cite{jia2020single}        &     28.71 &     77.29 &     18.83 \\
        ViT~\cite{huang2022adaptive}       &     20.86 &     85.05 &     43.89 \\
        SAFAS~\cite{sun2023rethinking}     &     33.40 &     70.89 &      8.92 \\
        FLIP-MCL~\cite{srivatsan2023flip}  &     18.82 &     86.94 &     46.48 \\    
        CFPL~\cite{liu2024cfpl}            & \B{16.15} & \B{88.65} & \B{53.98} \\
        InstructFLIP                       & \R{12.68} & \R{93.68} & \R{65.23} \\
        \bottomrule
        \end{tabular}
    }
}
\end{minipage}
\vspace{-2pt}
\end{table*}

\subsection{Content Feature-assist Query Fusion}
To combine the content and style information for spoof detection, we comprise the learned queries from each branch using the attention mechanisms presented in (\ref{eq:MSA}) and (\ref{eq:MCA}). Precisely, we reuse MSA and MCA operations to integrate the content and style queries along with the content feature $f_c$, creating a unified representation for spoof detection. The integration process is defined as
\begin{equation}
   \hat{Q} = Q+\psi(Q,\tilde{f}_c),
   \label{eq:concat_query}
\end{equation}
where $Q = Q_c \oplus Q_s \in \mathbb{R}^{2k\times d}$ is the concatenated representation of queries from content and style branches. Note that we do not include the style feature $f_s$ in the fusion process, as it is a domain-related representation and could potentially conflict with our goal of achieving robust generalization in the unified model.

The resulting unified representation $\hat{Q}$ from (\ref{eq:concat_query}) is divided into two segments. The first part contains a single foremost token, denoted as $t_{cls}$, which serves as the primary representation for the face spoofing classifier $\mathcal{C}$. The second part, $t_{cue}$, includes the remaining tokens, which capture auxiliary components and are fed into the cue generator $\mathcal{G}$ designed to produce attack hints to enhance the robustness. Given the uncertain nature of attack cues and the unavailable annotations within the dataset, we take inspiration from one-class training paradigms~\cite{huang2024oneclass}. Specifically, we apply a convolution layer that introduces guided, normalized noise, encouraging the model to generate a uniformly white map for fake samples, which signifies the presence of an attack on the input face. Conversely, the model is trained to produce an entirely blank map for real samples, indicating the absence of spoofing to assist in addressing FAS.

\subsection{Objectives}
Our proposed InstructFLIP is trained with the loss function that combines multiple components, which can be represented as
\begin{equation}
    \mathcal{L} = \lambda_{1}\mathcal{L}_{c} + \lambda_{2} \mathcal{L}_{s} + \lambda_{3}\mathcal{L}_{cls} + \lambda_{4}\mathcal{L}_{cue},
\label{eq:total_loss}
\end{equation}
where $\lambda_i$ are the hyperparameters for balancing each object. The first two terms\lin{, $\mathcal{L}_{c}$ and $\mathcal{L}_{s}$,} are introduced in content and style branches for carrying out instruction tuning. The third objective, $\mathcal{L}_{cls}$ optimizes the prediction for the face spoofing classifier $\mathcal{C}$ by cross-entropy. The last term\lin{, $\mathcal{L}_{cue}$,} is designed for learning the cue map from $\mathcal{G}$, which is defined as
\begin{equation}
\mathcal{L}_{cue} =     
   \begin{cases}
       \mathbf{d}^2 / \beta  &\quad \mathrm{if}~ \mathbf{d} < \beta   \;\\
       \mathbf{d} - \beta/2 &\quad \mathrm{otherwise}, \\ 
   \end{cases}
\label{eq:cue_loss}
\end{equation}
where $\mathbf{d} = | p_{cue} - y_{mask} |$ denotes the distance between the predicted cue map and GT binary mask. $\beta$ is the threshold at which to change between L1 and L2 loss.

\section{Experiment}

\subsection{Experimental Setup}
\noindent \textbf{Datasets and Evaluation Protocol.} \label{sec:dataset} To rigorously evaluate the proposed framework, we adopt a unified evaluation protocol that departs from traditional leave-one-out or one-to-one strategies~\cite{huang2022adaptive, sun2023rethinking, srivatsan2023flip, liu2024cfpl}, aiming to train a single model with strong cross-domain generalization while avoiding training redundancy. The CelebA-Spoof (CA)~\cite{2020ZhangCelebA-Spoof} is used for training, offering rich attribute-level annotations that enable the generation of diverse instruction-based image–text pairs. Specifically, eleven fine-grained spoof labels are used to construct textual supervisions for the content branch, while three imaging conditions, illumination, environment, and camera quality, serve as style-based contexts based on the extensive metadata available in the CA. Generalization performance is evaluated on seven publicly available FAS benchmarks: MSU-MFSD (M)\cite{wen2015face}, CASIA-FASD (C)\cite{zhang2012face}, Replay-Attack (I)\cite{chingovska2012effectiveness}, OULU-NPU (O)\cite{boulkenafet2017oulu}, WMCA (W)\cite{george2019biometric}, CASIA-CeFA (C)\cite{liu2021casia, liu2021cross}, and CASIA-SURF (S)~\cite{zhang2020casia, zhang2019dataset}. Under this unified setting, the model is trained once and evaluated on all target datasets concurrently, thereby providing a comprehensive evaluation of its generalization performance.

\noindent \textbf{Implementation Details.} \label{sec:implementation_details} Face images are aligned by MTCNN~\cite{zhang2016joint} and resized to $224\times224\times3$. We adopt the ViT-B/16 from CLIP~\cite{radford2021learning} as the visual feature extractor $\mathbf{E}$ and use FLAN-T5 base~\cite{chung2024scaling} as the frozen LLMs. Optimization is performed using the AdamW optimizer ~\cite{Loshchilov2017DecoupledWD} in combination with the OneCycleLR learning rate scheduler~\cite{smith2019super}. The initial learning rate is set to $1\times10^{-6}$, with a peak learning rate of $5\times10^{-6}$ and a weight decay of $10^{-6}$. All models are trained end-to-end for 20 epochs on 1 NVIDIA RTX 4090 GPU with a batch size of 24. The loss weights in InstructFLIP are set as follows: $\lambda_{1} = 0.4$, $\lambda_{2} = 0.4$, $\lambda_{3} = 0.15$, and $\lambda_{4} = 0.05$.

\noindent \textbf{Evaluation Metrics.} Following prior work~\cite{huang2022adaptive, sun2023rethinking, srivatsan2023flip, liu2024cfpl}, we evaluate model performance using three standard metrics: Half Total Error Rate (HTER), Area Under the Receiver Operating Characteristic Curve (AUC), and True Positive Rate (TPR) at a fixed False Positive Rate (FPR). To ensure fair and consistent evaluation, each model is trained five times with different random seeds, and the mean performance across runs is reported for all metrics.

\begin{table}[t]
\centering
\caption{Ablation study for the effectiveness of each component in InstructFLIP, where CB, SB, and Cue denote content branch, style branch, and cue generator, respectively.}
\label{tab:ablation_branches}
\scalebox{0.95}{
    \begin{tabular}{ccc|ccc}
    \toprule
    CB & SB & Cue & HTER$\downarrow$ & AUC$\uparrow$ & TPR$\uparrow$@FPR=1\% \\
    \midrule
         --      &       --     &       --     & 21.96~({\footnotesize +0.0\%}) & 82.88~({\footnotesize +0.0\%}) & 26.89~({\footnotesize +0.0\%})  \\
    $\checkmark$ &       --     &       --     & 14.25~({\footnotesize +35.1\%}) & 90.23~({\footnotesize +8.9\%}) & 32.32~({\footnotesize +20.2\%})  \\
         --      & $\checkmark$ &       --     & 16.25~({\footnotesize +26.0\%}) & 88.27~({\footnotesize +6.5\%}) & 41.92~({\footnotesize +55.9\%})  \\
    $\checkmark$ & $\checkmark$ &       --     & 13.25~({\footnotesize +39.7\%}) & 91.21~({\footnotesize +10.1\%}) & 60.84~({\footnotesize +126.3\%})  \\
    $\checkmark$ & $\checkmark$ & $\checkmark$ & \textbf{12.68~({\footnotesize +42.3\%})} & \textbf{93.68~({\footnotesize +13.0\%})} & \textbf{65.23~({\footnotesize +142.6\%})}  \\
    \bottomrule
    \end{tabular}
}
\end{table}

\subsection{Unified FAS Performance}

\noindent \textbf{Baseline Methods}. The primary and most competitive baselines for evaluating the proposed InstructFLIP method include SSDG~\cite{jia2020single}, ViT~\cite{huang2022adaptive}, SAFAS~\cite{sun2023rethinking}, FLIP~\cite{srivatsan2023flip}, CFPL~\cite{liu2024cfpl}, BUDoPT~\cite{liu2024bottom}, DiffFAS~\cite{ge2024difffas}, and FGPL~\cite{hu2024fine}. While ViT provides results for both zero-shot and few-shot settings, we report only the zero-shot setting to maintain consistency with our unified protocol, and we exclude the adaptive ViTAF variant specific to the few-shot scenario. Furthermore, due to the unavailability of publicly released codes for BUDoPT, DiffFAS, and FGPL, these methods are excluded from our experimental comparisons. Consequently, we reimplemented SSDG, ViT, SAFAS, FLIP, and CFPL under our unified evaluation framework and presented their results in the subsequent subsections.

\noindent \textbf{Comparative Results}. \label{sec:unified_result} Tables~\ref{tab:unified_a} and~\ref{tab:unified_b} present comparative results between existing FAS methods and the proposed InstructFLIP on the proposed unified evaluation protocol. InstructFLIP consistently outperforms all prior approaches across all evaluation metrics and datasets, demonstrating its strong robustness and adaptability. In terms of HTER, InstructFLIP achieves significant reductions compared to the current SOTA method CFPL~\cite{liu2024cfpl}, with improvements of 37$\%$, 47$\%$, 3.5$\%$, 16$\%$, 28$\%$, and 25$\%$ on the M, C, I, O, W, and S datasets, respectively. Beyond HTER, consistent gains in AUC reveal InstructFLIP’s ability to learn robust and discriminative features for distinguishing genuine from spoofed faces. Improvements in TPR@FPR=1$\%$ further demonstrate the effectiveness of InstructFLIP in maintaining a low false positive rate while ensuring high detection accuracy, which is essential for practical deployment where both security and user experience are critical.

Despite the overall strong performance, the results on the CASIA-CeFA (C) dataset remain relatively modest, indicating potential limitations in modeling granular cues. This suggests a direction for future work focused on enhancing the model’s sensitivity to subtle cultural and environmental variations through more expressive and adaptive instruction-tuning strategies. Overall, the results summarized in Table~\ref{tab:unified_c} validate that InstructFLIP offers a well-balanced and generalizable solution for robust FAS across diverse domains.

\subsection{Ablation Study} 
In this subsection, we perform a series of ablation studies to assess the effectiveness of the proposed method from multiple perspectives. All ablation experiments are conducted using the meta-domain for training, and the reported results represent the average performance across the seven target datasets, following the unified evaluation protocol described in Section~\ref{sec:dataset}.

\noindent \textbf{Effectiveness of proposed components.} We begin by evaluating the contribution of each component in the proposed InstructFLIP framework through an ablation study, including the content branch (CB), style branch (SB), and cue generation (Cue). The baseline model is constructed by removing all proposed components and relying solely on the given image, following the conventional FAS setup for a fair comparison. The comparative results, summarized in Table~\ref{tab:ablation_branches}, illustrate the impact of each module on overall performance.

\begin{table}[t]
\centering
\caption{Ablation study to assess the effectiveness of frozen LLMs in InstructFLIP. InstructFLIP$^\dagger$ replaces LLM in the dual branches with classification heads.}
\label{tab:ab_llms}
\scalebox{1}{
    \begin{tabular}{l|ccc}
        \toprule
        Method                      & HTER$\downarrow$ & AUC$\uparrow$  & TPR$\uparrow$@FPR=1\% \\
        \midrule
        CFPL~\cite{liu2024cfpl}     & 16.15            & 88.65          & 53.98                 \\
        InsturctFLIP$^\dagger$      & 14.39            & 89.97          & 48.33                 \\
        InstructFLIP                & \textbf{12.68}   & \textbf{93.68} & \textbf{65.23}        \\
        \bottomrule
    \end{tabular}
}
\vspace{6pt}
\end{table}
\begin{table}[t]
\centering
\caption{Ablation studies on effectiveness of category diversity. In the table, $Pr.$, $Re.$, $M_{2D}$, and $M_{3D}$ represent print, replay, 2D mask, and 3D mask, respectively.}
\label{tab:ablation_dataset}
\setlength{\tabcolsep}{3pt} 
\scalebox{1}{
    \begin{tabular}{ccccc|ccc}
    \toprule
    Real & $Pr.$ & $Re.$ & $M_{2D}$ & $M_{3D}$ & ~HTER$\downarrow$~ & ~AUC$\uparrow$~ & TPR$\uparrow$@FPR=$1\%$~ \\
    \midrule
    1 & 1 & 1 & 1 & 1 & 19.11 & 87.14 & 39.13  \\
    1 & 2 & 2 & 2 & 1 & 18.57 & 87.3 & 38.3  \\
    1 & 3 & 3 & 3 & 1 & \textbf{12.68} & \textbf{93.68} & \textbf{65.23}  \\
    \bottomrule
    \end{tabular}
}
\end{table}



Introducing the CB to the baseline model resulted in substantial improvements: +35.1$\%$ in HTER, +8.9$\%$ in AUC, and +20.2$\%$ in TPR@FPR=1$\%$. These results demonstrate the critical role of CB, which utilizes fine-grained labels to enhance the model's ability to effectively capture features that differentiate genuine inputs from spoofing attacks. The contribution of the SB is also evaluated, which further improved performance by +26.0$\%$ in HTER, +6.5$\%$ in AUC, and +55.9$\%$ in TPR@FPR=1$\%$. The considerable boost in TPR@FPR=1$\%$ indicates that SB effectively models non-spoofing patterns, thereby reducing overfitting and improving generalization. When both CB and SB were integrated, the model achieved even greater performance gains: +39.7$\%$ in HTER, +10.1$\%$ in AUC, and +126.3$\%$ in TPR@FPR=1$\%$, demonstrating their complementary nature in capturing both spoof-relevant and irrelevantly contextual features. Furthermore, the addition of Cue led to SOTA performance, with improvements of +42.3$\%$ in HTER, +13.0$\%$ in AUC, and an impressive +142.6$\%$ increase in TPR@FPR=1$\%$. The module enriches the fused representation by providing auxiliary guidance, helping the model to more effectively differentiate between genuine and spoofed samples. These results confirm the effectiveness of each proposed component and the synergistic benefits of their integration in enhancing robustness and generalization.

\noindent \textbf{Effectiveness of LLMs adaptation.} To better understand whether the performance gains observed in Section~\ref{sec:unified_result} are primarily attributed to the use of frozen large language models (LLMs) or to the incorporation of fine-grained labels from the meta-domain, we conduct an additional experiment with an ablated variant, InstructFLIP$^\dagger$. This variant retains the content and style branches but replaces the frozen LLMs with lightweight classification heads. As shown in Table~\ref{tab:ab_llms}, incorporating fine-grained semantic signals already yields performance superior to CFPL, and further integrating LLMs leads to the best overall results. These results point out the complementary roles of structured supervision and language-based reasoning in enhancing model generalization and discriminative capability.

\begin{table}[t]
\centering
\caption{Ablation studies on effectiveness of style prompts.}
\label{tab:ab_style}
\scalebox{0.91}{
    \begin{tabular}{c|c|c|c}
    \toprule
    Style prompt                      & HTER$\downarrow$ & AUC$\uparrow$  & TPR$\uparrow$@FPR=1\% \\
    \midrule
    Style1                            & 20.65            & 85.60          & 37.48                 \\
    Style1 + Style2                   & 19.25            & 86.71          & 45.08                 \\
    Style1 + Style2 + Style3          & \textbf{12.68}   & \textbf{93.68} & \textbf{65.23}        \\
    \bottomrule
    \end{tabular}
}
\end{table}

\noindent \textbf{Effectiveness of category diversity.} To evaluate the effect of semantic granularity in spoof category annotations, we conduct an ablation study on the category diversity used during training. The results, shown in Table~\ref{tab:ablation_dataset}, demonstrate the positive impact of incorporating detailed subcategories for print, replay, and 2D mask attacks. As the number of fine-grained classes increases from 1 to 3, the model exhibits consistent improvements across all key evaluation metrics. Specifically, the HTER decreases from 19.11 to 12.68, indicating a substantial reduction in overall error. The AUC improves from 87.14 to 93.68, suggesting enhanced discriminative capability, while the TPR@FPR=1$\%$ increases significantly from 39.13 to 65.23, highlighting improved sensitivity under low false positive constraints. These results indicate that richer semantic supervision allows the model to better capture intra-class variations, thereby improving robustness and reliability in real-world use.

\noindent \textbf{Effectiveness of style prompts.} In order to investigate the impact of style prompt diversity on model performance, we conduct an ablation study by progressively incorporating different style prompts into InstructFLIP, with the results summarized in Table~\ref{tab:ab_style}. Starting to use only a single style prompt related to illumination conditions (style1) yields an HTER of 20.65, an AUC of 85.60, and a TPR@FPR=1$\%$ of 37.48. Adding an additional prompt reflecting environmental context (style1 + style2) leads to noticeable improvements, reducing the HTER to 19.25, increasing the AUC to 86.71, and raising the TPR@FPR=1$\%$ to 45.08. Further incorporating a third prompt associated with camera quality (style1 + style2 + style3) significantly enhances SOTA performance. The results demonstrate the importance of incorporating diverse style prompts to model non-spoof-related variations. By enriching the instruction space, the model becomes more effective at disentangling spoof-relevant and confounding factors, leading to improved generalization and robustness across diverse domain conditions.


\begin{table}[t]
\centering
\caption{Question inputs for VLMs.}
\label{tab:question_inputs}
\scalebox{0.91}{
    \begin{tabular}{c|p{7.2cm}}
    \toprule
    Type & \multicolumn{1}{c}{Question inputs} \\
    \midrule
    \midrule
    \multirow{5}{*}{Content} & Which type of spoof is in this image? \newline (1) Real face (2) Photo (3) Poster (4) A4-paper (5) 2D face mask (6) 2D upper-body mask (7) 2D region mask (8) PC screen (9) Pad screen (10) Phone screen (11) 3D mask \\ 
    \midrule
    \multirow{2}{*}{Style1} & What is the illumination condition in this image? \newline (1) Normal (2) Strong (3) Back (4) Dark \\ 
    \midrule
    \multirow{2}{*}{Style2} & What is the environment in this image? \newline (1) Indoor (2) Outdoor \\ 
    \midrule
    \multirow{2}{*}{Style3} & What is the camera quality in this image? \newline (1) Low (2) Medium (3) High \\ 
    \bottomrule
    \end{tabular}
}
\vspace{-4pt}
\end{table}

\subsection{Qualitative Results}
In this subsection, we present qualitative results to illustrate the strengths and limitations of the proposed method. The analysis includes successful cases, failure cases, and comparisons with open-source VLMs to provide a clearer understanding of InstructFLIP’s capabilities. To ensure a fair and informative comparison, each input to the VLMs is structured using four prompt components—content, style1, style2, and style3—as defined in Table~\ref{tab:question_inputs}. This design enables the models to develop a comprehensive understanding of each sample’s semantic and contextual attributes.

\begin{table*}[t]
\centering
\caption{Illustration of (a) True-positive, (b) True-negative, (c) False-positive, and (d) False-negative samples predicted by the proposed method. \R{Red} indicates incorrect answers.}
\label{tab:visualization}
\begin{minipage}[t]{0.49\textwidth}
\centering
\subfloat[\textbf{True-positive sample}]{
    \label{tab:visualization_a}
    \scalebox{0.78}{
        \begin{tabular}{@{}c|c|c|c|c|c@{}}
        \toprule
        Raw image & Cue map & Fake score & Q-type & Answer & GT \\
        \midrule
        \multirow{4}{*}{\includegraphics[width=0.175\textwidth]{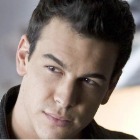}} 
        & \multirow{4}{*}{\includegraphics[width=0.175\textwidth]{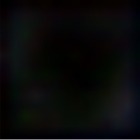}} 
        & \multirow{4}{*}{0.001} 
        & Content & (1) Real face & (1) Real face \\
        & & & Style1 & (3) Back & (3) Back \\
        & & & Style2 & (2) Outdoor & (2) Outdoor \\
        & & & Style3 & (3) High & (3) High \\
        \bottomrule
        \end{tabular}
    }
}
\end{minipage}
\begin{minipage}[t]{0.49\textwidth}
\centering
\subfloat[\textbf{True-negative sample}]{
    \label{tab:visualization_b}
    \scalebox{0.78}{
        \begin{tabular}{@{}c|c|c|c|c|c@{}}
        \toprule
        Raw image & Cue map & Fake score & Q-type & Answer & GT \\
        \midrule
        \multirow{4}{*}{\includegraphics[width=0.175\textwidth]{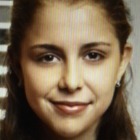}} 
        & \multirow{4}{*}{\includegraphics[width=0.175\textwidth]{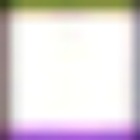}} 
        & \multirow{4}{*}{0.998} & Content & (8) PC Screen & (8) PC Screen \\
        & & & Style1 & (1) Normal & (1) Normal \\
        & & & Style2 & (1) Indoor & (1) Indoor \\
        & & & Style3 & (2) Medium & (2) Medium \\
        \bottomrule
        \end{tabular}
    }
}
\end{minipage}

\begin{minipage}[t]{0.49\textwidth}
\centering
\subfloat[\textbf{False-positive sample}]{
    \label{tab:visualization_c}
    \scalebox{0.78}{
        \begin{tabular}{@{}c|c|c|c|c|c@{}}
        \toprule
        Raw image & Cue map & Fake score & Q-type & Answer & GT \\
        \midrule
        \multirow{4}{*}{\includegraphics[width=0.175\textwidth]{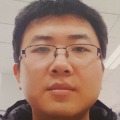}} 
        & \multirow{4}{*}{\includegraphics[width=0.175\textwidth]{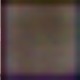}} 
        & \multirow{4}{*}{0.055} 
        & Content & \R{(1) Real face} & (3) Poster \\
        & & & Style1 & \R{(3) Back} & (1) Normal \\
        & & & Style2 & (1) Indoor & (1) Indoor \\
        & & & Style3 & (2) Medium & (2) Medium \\
        \bottomrule
        \end{tabular}
    }
}
\end{minipage}
\begin{minipage}[t]{0.49\textwidth}
\centering
\subfloat[\textbf{False-negative sample}]{
    \label{tab:visualization_d}
    \scalebox{0.78}{
        \begin{tabular}{@{}c|c|c|c|c|c@{}}
        \toprule
        Raw image & Cue map & Fake score & Q-type & Answer & GT \\
        \midrule
        \multirow{4}{*}{\includegraphics[width=0.175\textwidth]{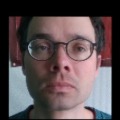}} 
        & \multirow{4}{*}{\includegraphics[width=0.175\textwidth]{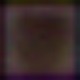}} 
        & \multirow{4}{*}{0.851} 
        & Content & \R{(8) PC screen} & (1) Real face \\
        & & & Style1 & (3) Back & (3) Back \\
        & & & Style2 & (1) Indoor & (1) Indoor \\
        & & & Style3 & \R{(3) High} & (2) Medium \\
        \bottomrule
        \end{tabular}
    }
}
\end{minipage}
\end{table*}
\begin{table*}[t]
\centering
\caption{Comparison of model responses. \R{Red} indicates incorrect answers and \DF{gray} represents indirect or ambiguous responses.}
\label{tab:sup_comparison_llms}
\vspace{-4pt}
\scalebox{0.78}{
    \begin{tabular}{c|c|p{1.9cm}|p{10cm}|p{1.9cm}|p{1.9cm}}
    \toprule
    \multirow{2}{*}{\textbf{Raw image}} & \multicolumn{1}{c|}{\multirow{2}{*}{\textbf{Question type}}} & \multicolumn{3}{c|}{\textbf{Response}} & \multicolumn{1}{c}{\multirow{2}{*}{\textbf{GT}}} \\
    \cline{3-5} 
    & & \multicolumn{1}{c|}{\raisebox{-0.5ex}{InstructBLIP}} & \multicolumn{1}{c|}{\raisebox{-0.5ex}{ChatGPT-4o}} & \multicolumn{1}{c|}{\raisebox{-0.5ex}{InstructFLIP}} & \\
    \midrule
    \midrule
    \multirow{2}{*}{\includegraphics[width=0.13\textwidth]{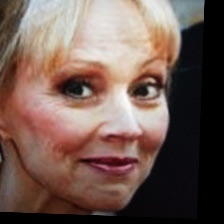}} & \multirow{3}{*}{Content} & \multirow{3}{*}{\R{(1) Real face}} & \R{I cannot determine spoof types from an uploaded image. However, you can analyze this image using algorithms trained for Face Anti-Spoofing tasks. Let me know if you need guidance on setting up such an analysis.} & \multirow{3}{*}{(9) Pad screen} & \multirow{3}{*}{(9) Pad screen} \\
    \cline{2-6}
    & \multirow{3}{*}{Illumination} & \multirow{3}{*}{(1) Normal} & \R{\raisebox{-0.5ex}{I cannot visually analyze the illumination condition from the image. However,} \raisebox{-0.5ex}{if you have an algorithm to evaluate the illumination or would like to process} \raisebox{-0.5ex}{the image computationally, I can guide you through that.}} & \multirow{3}{*}{(1) Normal} & \multirow{3}{*}{(1) Normal} \\
    \midrule
    \multirow{2}{*}{\includegraphics[width=0.13\textwidth]{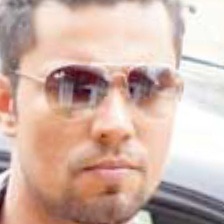}} & \multirow{3}{*}{Content} & \multirow{3}{*}{(1) Real face} & \R{I cannot determine spoof types from an uploaded image. However, you can analyze this image using algorithms trained for Face Anti-Spoofing tasks. Let me know if you need guidance on setting up such an analysis.} & \multirow{3}{*}{(1) Real face} & \multirow{3}{*}{(1) Real face} \\
    \cline{2-6}
    & \multirow{3}{*}{Environment} & \multirow{3}{*}{\R{(1) Indoor}} & \DF{\raisebox{-0.5ex}{The image appears to be taken in an outdoor environment, as suggested by} \raisebox{-0.5ex}{the presence of natural sunlight, reflections in the sunglasses, and background} \raisebox{-0.5ex}{elements such as trees or open space.}} & \multirow{3}{*}{(2) Outdoor} & \multirow{3}{*}{(2) Outdoor} \\
    \bottomrule
    \end{tabular}
}
\end{table*}

\noindent \textbf{Illustration of successful cases.} Tables~\ref{tab:visualization_a} and~\ref{tab:visualization_b} present successful cases processed by the proposed InstructFLIP model. For the true positive sample in Table~\ref{tab:visualization_a}, the model assigns a fake score of 0.001, indicating a negligible likelihood of misclassifying the live face as a spoof. In contrast, the true-negative sample in Table~\ref{tab:visualization_b} receives a fake score of 0.998, reflecting high confidence in correctly identifying the spoof. The LLM-generated predictions—including spoof type, illumination, environment, and camera quality—demonstrate InstructFLIP’s ability to generalize to unseen domains by leveraging structured instructional inputs from the meta domain. Furthermore, the cue representations contribute significantly to differentiating between live and spoofed samples, highlighting their importance in the overall decision-making process.

\noindent \textbf{Illustration of failure cases.} To better understand the limitations of our model, we analyzed representative failure cases, categorizing them into false positives and false negatives. For the false-positive case illustrated in Table~\ref{tab:visualization_c}, the model misclassified a spoofed poster as a real face, likely due to difficulty in distinguishing subtle material cues like gloss and texture. It also misjudged backlighting as Normal, indicating sensitivity to slight luminance variations, though it accurately predicted other style attributes such as environment and camera quality. For the false-negative case shown in Table~\ref{tab:visualization_d}, a real face was classified as a spoofed PC screen, possibly due to overfitting to reflective surface patterns. While illumination was correctly predicted, the camera quality was overestimated, revealing inconsistency in assessing image sharpness.

\noindent \textbf{Comparison with open VLMs.} We qualitatively compare our InstructFLIP with InstructBLIP and GPT-4o using both content and style-based instructions on a fake and a live images, respectively, as shown in Table~\ref{tab:sup_comparison_llms}. For content-based instructions, InstructFLIP consistently outperforms the other VLMs, accurately identifying fake samples as Pad screens and correctly recognizing live faces. In contrast, InstructBLIP frequently misclassifies spoofed faces as real, indicating a limited capacity to capture spoofing cues. GPT-4o, although capable of generating general explanations, refrains from making explicit predictions and instead suggests computational techniques, reducing its utility in this task. For style-based instructions, InstructFLIP provides strong performance across diverse conditions, accurately predicting illumination, camera quality, and environment. While InstructBLIP performs reasonably on simpler attributes like illumination, it struggles with more nuanced aspects such as environment, occasionally misclassifying Outdoor as Indoor. GPT-4o avoids direct responses in the style setting, revealing a lack of task-specific grounding. These findings show InstructFLIP’s adaptability across both instruction types and underscore the importance of efficient contextual understanding in VLMs.

\section{Conclusion}
We present InstructFLIP, a novel instruction approach for FAS, featuring a unified architecture that decouples content and style representations. By leveraging a dual-branch design, InstructFLIP effectively captures spoof-related patterns from a meta domain, ensuring robust generalization across diverse domains. We employ the Q-Former to encode semantic information from content and style prompts, allowing the model to comprehensively understand diverse spoofing types and environmental conditions. We also propose a generalized query fusion strategy and additional cue maps that enhance FAS performance by efficiently capturing domain-invariant characteristics. Experimental results validate InstructFLIP’s efficacy in achieving strong generalization,  marking a significant step toward practical, adaptable FAS solutions in real-world applications. Future work may explore extending this framework to other visual tasks where robustness across domains remains challenging, further advancing instruction-driven generalization.

\section{Acknowledgments}
\noindent This work is partially supported by the National Science and Technology Council, Taiwan, under Grant: NSTC-112-2628-E-002-033-MY4, and was financially supported in part by the Center of Data Intelligence: Technologies, Applications, and Systems, National Taiwan University (Grants: 114L900901/114L900902/114L900903), from the Featured Areas Research Center Program within the framework of the Higher Education Sprout Project by the Ministry of Education, Taiwan.

\bibliographystyle{ACM-Reference-Format}
\balance
\bibliography{ref}

\end{document}


\title{Supplementary Materials for InstructFLIP: Exploring Unified Vision-Language Model for Face Anti-spoofing}

\maketitle

We present additional experiments and analyses to thoroughly assess the effectiveness of the proposed InstructFLIP framework. Section~\ref{sec:sup_loss} conducts a detailed ablation study on the objective function by varying the loss weight configurations. Section~\ref{sec:sup_llm_size} examines how the size of the frozen LLM affects performance, while Section~\ref{sec:sup_compare_to_ifas} compares InstructFLIP against the I-FAS baseline. Section~\ref{sec:sup_cost_of_instructflip} analyzes the computational cost of our approach. Section~\ref{sec:sup_no_rich_labels} evaluates the model’s applicability beyond CelebA-Spoof. Lastly, Section~\ref{sec:sup_qformer_layers} explores the impact of Q-Former depth on model performance.

\renewcommand{\thetable}{S\arabic{table}}
\begin{table}[h]
\centering
\caption{Ablation study for the effectiveness of weights ($\lambda_{1}, \lambda_{2}, \lambda_{3}, \lambda_{4}$) in loss function. The last row represents the optimal configuration in our experiments.}
\label{tab:sup_loss_weight}
\vspace{-4pt}
\scalebox{1}{
    \begin{tabular}{c|c|c|c}
        \toprule
        \multirow{2}{*}{$\lambda_{1},~\lambda_{2},~\lambda_{3},~\lambda_{4}$} & 
        \multirow{2}{*}{HTER$\downarrow$} &
        \multirow{2}{*}{AUC$\uparrow$} & 
        TPR$\uparrow$@ \\
        &&& FPR=1\% \\
        \midrule
                 0.1,     \DF{0.4},    \DF{0.15},    \DF{0.05} &           15.61  &           89.74  &           52.67      \\
                 0.2,     \DF{0.4},    \DF{0.15},    \DF{0.05} &           13.84  &           90.36  &           54.17      \\
                 0.3,     \DF{0.4},    \DF{0.15},    \DF{0.05} &           13.25  &           91.33  &           59.25      \\
                 0.5,     \DF{0.4},    \DF{0.15},    \DF{0.05} &           15.23  &           89.38  &           52.70      \\
        \midrule
            \DF{0.4},          0.1,    \DF{0.15},    \DF{0.05} &           14.42  &           89.88  &           54.83      \\
            \DF{0.4},          0.2,    \DF{0.15},    \DF{0.05} &           15.49  &           89.13  &           55.20      \\
            \DF{0.4},          0.3,    \DF{0.15},    \DF{0.05} &           13.48  &           92.85  &           62.48      \\
            \DF{0.4},          0.5,    \DF{0.15},    \DF{0.05} &           15.90  &           87.93  &           53.11      \\
        \midrule
            \DF{0.4},     \DF{0.4},         0.05,    \DF{0.05} &           14.81  &           89.40  &           51.31      \\
            \DF{0.4},     \DF{0.4},          0.1,    \DF{0.05} &           14.20  &           90.94  &           57.23      \\
            \DF{0.4},     \DF{0.4},          0.2,    \DF{0.05} &           13.90  &           91.00  &           60.17      \\
            \DF{0.4},     \DF{0.4},         0.25,    \DF{0.05} &           14.70  &           89.95  &           53.31      \\
        \midrule
            \DF{0.4},     \DF{0.4},    \DF{0.15},         0.01 &           14.78  &           89.69  &           58.76      \\
            \DF{0.4},     \DF{0.4},    \DF{0.15},         0.03 &           13.58  &           90.68  &           59.25      \\
            \DF{0.4},     \DF{0.4},    \DF{0.15},         0.07 &           13.49  &           90.97  &           61.35      \\
            \DF{0.4},     \DF{0.4},    \DF{0.15},         0.09 &           14.57  &           89.70  &           53.79      \\
        \midrule
                 0.4,          0.4,         0.15,         0.05 &   \textbf{12.68} &   \textbf{93.68} &   \textbf{65.23}     \\
        \bottomrule
    \end{tabular}
}
\end{table}

\section{Detailed Ablation Study for Objectives}
\label{sec:sup_loss}
In addition to the ablation study on $\lambda_{1}$ and $\lambda_{3}$ presented in the main paper, we further extend our analysis to include the remaining loss weight parameters. Table~\ref{tab:sup_loss_weight} reports the impact of varying each $\lambda_i$ independently, with all other weights fixed at their default values (highlighted in gray for reference). Overall, the model exhibits relatively low sensitivity to most weight changes. Consistent with the findings in the main paper, $\lambda_{1}$ exerts the most significant influence, with even minor adjustments causing notable fluctuations in performance. $\lambda_{2}$ exhibits slight instability in the loss function due to the diverse nature of style descriptions, which introduces variations in the learning process. In contrast, $\lambda_{3}$, and $\lambda_{4}$ display relatively stable effects, contributing to more balanced results overall. The optimal configuration presented in the last row yields the best performance, emphasizing the critical role of precise weight tuning in enhancing our model's effectiveness.

\renewcommand{\thetable}{S\arabic{table}}
\begin{table}[t]
\centering
\vspace{4pt}
\caption{Ablation study on the effectiveness of different frozen LLM sizes. FLAN-T5-base is selected as the default LLM in InstructFLIP due to its favorable balance of accuracy and computational efficiency.}
\label{tab:sup_llm_size}
\vspace{-4pt}
\scalebox{1}{
    \begin{tabular}{l|c|c|c|c}
        \toprule
        \multirow{2}{*}{Model} &
        \multirow{2}{*}{Params} &
        \multirow{2}{*}{HTER$\downarrow$} &
        \multirow{2}{*}{AUC$\uparrow$} & 
        TPR$\uparrow$@ \\
        &&&& FPR=1\% \\
        \midrule
        FLAN-T5-small   &  80M &        14.66  &         90.11  &         54.82  \\
        FLAN-T5-base    & 250M &        12.68  & \textbf{93.68} &         65.23  \\
        FLAN-T5-large   & 750M &\textbf{12.24} &         92.35  & \textbf{65.33} \\
        FLAN-T5-xl      &   3B &        12.62  &         92.52  &         63.13  \\
        \bottomrule
    \end{tabular}
}
\end{table}

\section{Ablation Study on Frozen LLM Sizes}
\label{sec:sup_llm_size}
To investigate the impact of LLM size on proposed InstructFLIP performance, we conduct an ablation study using four frozen FLAN-T5 variants: small (80M), base (250M), large (750M), and XL (3B). As shown in Table~\ref{tab:sup_llm_size}, increasing the model size generally improves performance across most metrics. FLAN-T5-large achieves the lowest HTER of 12.24 and the highest TPR@FPR=1\% of 65.33, indicating strong discriminative capability and robustness to false positives. Interestingly, FLAN-T5-base outperforms other variants in AUC of 93.68, suggesting that mid-sized models may strike a better balance between representation capacity and calibration. While FLAN-T5-xl has the largest number of parameters, it underperforms compared to FLAN-T5-large. This degradation is attributed to hardware constraints, specifically the use of an NVIDIA RTX 4090 GPU with 24GB of memory, which necessitated significantly smaller batch sizes. We also observed that such limited batch sizes impair optimization stability and convergence, even for smaller LLMs. Considering the trade-off between model complexity, computational efficiency, and overall performance, we adopt FLAN-T5-base as the default LLM in InstructFLIP due to its balanced performance and manageable resource requirements.

\renewcommand{\thetable}{S\arabic{table}}
\begin{table*}[t]
\centering
\caption{Comparison of I-FAS and InstructFLIP under the proposed protocol on MCIOW datasets.}
\label{tab:sup_ifas}
\vspace{-4pt}
\setlength{\tabcolsep}{3pt}
\scalebox{0.95}[0.95]{
    \begin{tabular}{lllcclcclcclcclcclccl|c}
    \toprule
    \multirow{3}{*}{\BT{Method}} & \multirow{3}{*}{\BT{Venue}} & &
    \multicolumn{2}{c}{\BT{M}} & &
    \multicolumn{2}{c}{\BT{C}} & &
    \multicolumn{2}{c}{\BT{I}} & &
    \multicolumn{2}{c}{\BT{O}} & &
    \multicolumn{2}{c}{\BT{W}} & &
    \multicolumn{2}{c}{\BT{Avg.}} \\
    \cmidrule{4-5} \cmidrule{7-8} \cmidrule{10-11} \cmidrule{13-14} \cmidrule{16-17} \cmidrule{19-20}
    & & &
    ~HTER$\downarrow$~ & ~AUC$\uparrow$~ & &
    ~HTER$\downarrow$~ & ~AUC$\uparrow$~ & &
    ~HTER$\downarrow$~ & ~AUC$\uparrow$~ & &
    ~HTER$\downarrow$~ & ~AUC$\uparrow$~ & & 
    ~HTER$\downarrow$~ & ~AUC$\uparrow$~ & &
    ~HTER$\downarrow$~ & ~AUC$\uparrow$~ \\ 
    \midrule
    I-FAS                & AAAI'25 &&     5.63  & \BT{98.73} && \BT{1.11} & \BT{99.88} &&     9.15  &     95.12  &&     14.86  &     91.68  &&     20.07  &     89.17  &&    12.70   &     94.92  \\
    InstructFLIP         &   -     && \BT{5.52} &     98.12  &&     1.47 &      99.79  && \BT{9.12} & \BT{96.17} && \BT{14.33} & \BT{94.79} && \BT{19.51} & \BT{89.90} && \BT{12.49} & \BT{95.76}  \\
    \bottomrule
    \end{tabular}
}
\vspace{4pt}
\end{table*}

\section{Compare to I-FAS}
\label{sec:sup_compare_to_ifas}
For I-FAS~[44], we acknowledge this work and note that it employs a 1-to-11 protocol with less commonly used datasets in domain generalization-based FAS. Additionally, it does not provide an official code, which prevents us from including this model under the same protocol in the main table. Nevertheless, its evaluation partially aligns with ours on the MCIOW subset, where InstructFLIP demonstrates superior performance (see Table~\ref{tab:sup_ifas}). We will discuss and include these FAS methods in our comparison once the implementations are made public in the revision.

\renewcommand{\thetable}{S\arabic{table}}
\begin{table}[t]
\centering
\caption{Comparison of model size, computational cost, and inference speed across FAS methods. InstructFLIP$^{-}$ denotes the variant of InstructFLIP with the frozen LLM removed during inference. All values are reported in terms of parameter count (M), FLOPs (G), and average inference time (s) per sample with standard deviation on a single NVIDIA RTX 4090 GPU.}
\label{tab:sup_speed}
\scalebox{1}{
    \begin{tabular}{l|ccc}
        \toprule
        Method              & Params (M)  & FLOPs (G)  & Inference time (s)  \\
        \midrule
        SSDG~[3]            & 11.97       &  3.64      & 0.005 \\ 
        ViT~[37]            & 86.19       & 33.72      & 0.029 \\ 
        SAFAS~[8]           & 11.44       &  3.64      & 0.005 \\ 
        FLIP-MCL~[10]       & 149.88      & 33.71      & 0.030 \\ 
        CFPL~[11]           & 155.08      & 45.46      & 0.070 \\ 
        InstructFLIP$^{-}$  & 86.19       & 33,72      & 0.030 \\ 
        InstructFLIP        & 195.71      & 39.90      & 0.046 \\ 
        \bottomrule
    \end{tabular}
}
\end{table}
\section{Cost of InstructFLIP}
\label{sec:sup_cost_of_instructflip}
InstructFLIP incorporates an LLM during training, but it supports a lightweight variant, InstructFLIP$^{-}$, where the frozen LLM is removed at inference. As shown in Table~\ref{tab:sup_speed}, InstructFLIP$^{-}$ achieves an inference time of 0.030 ± 0.0011 seconds per sample—comparable to other real-time FAS methods such as FLIP-MCL (0.030 s) and ViT (0.029 s). Its computational cost (86.19M parameters and 33.72G FLOPs) is also significantly lower than CFPL (155.08M / 45.46G), making it well-suited for practical deployment.

\renewcommand{\thetable}{S\arabic{table}}
\begin{table}[t]
\centering
\setlength{\tabcolsep}{4pt} 
\caption{Evaluation under leave-one-out protocol on MCIO datasets. InstructFLIP$^{\ast}$ is a variant of InstructFLIP that excludes the style branch and is trained using binary yes-or-no questions only.} 
\label{tab:sup_leave_one_out}
\scalebox{1}{
    \begin{tabular}{llccc}
    \toprule
    Method            &  HTER$\downarrow$ &  AUC$\uparrow$ & TPR$\uparrow$@FPR=1\% \\
    \midrule
    CFPL~[11]          &             3.60  & \textbf{99.00} &                83.99  \\
    InstructFLIP$^{\ast}$  &     \textbf{3.04} &         98.45  &        \textbf{93.43} \\
    \bottomrule
    \end{tabular}
}
\end{table}


\section{Applicability beyond CelebA-Spoof}
\label{sec:sup_no_rich_labels}
To test InstructFLIP's applicability without style annotations, we introduce a simplified variant, InstructFLIP$^{\ast}$, which removes the style branch and uses binary yes/no questions (e.g., “Is this a real face?”). This variant follows CFPL’s setup and requires only RGB input. As shown in Table~\ref{tab:sup_leave_one_out}, InstructFLIP$^{\ast}$ still achieves competitive results, indicating that our framework generalizes well even without explicit style supervision. We will include these results in the supplementary to highlight this practical flexibility.

\renewcommand{\thetable}{S\arabic{table}}
\begin{table}[t]
\centering
\caption{Ablation study on the effectiveness of depth of Q-Former layers. The two-layer Q-Former achieves the best overall performance.}
\label{tab:sup_qformer_layers}
\vspace{-4pt}
\scalebox{1}{
    \begin{tabular}{c|c|c|c}
        \toprule
        \multirow{2}{*}{Depth} &
        \multirow{2}{*}{HTER$\downarrow$} &
        \multirow{2}{*}{AUC$\uparrow$} & 
        TPR$\uparrow$@ \\
        &&& FPR=1\% \\
        \midrule
        1 &         13.53  &         91.31  &         58.20  \\
        2 & \textbf{12.68} & \textbf{93.68} & \textbf{65.23} \\
        4 &         12.71  &         92.89  &         64.47  \\
        8 &         12.89  &         93.57  &         62.19  \\
        \bottomrule
    \end{tabular}
}
\end{table}

\section{Ablation Study on the Depth of Q-Former}
\label{sec:sup_qformer_layers}
We examines the effect of Q-Former depth on model performance by varying the number of layers in both the content and style branches. As shown in Table~\ref{tab:sup_qformer_layers}, the configuration with two Q-Former layers delivers the best overall results. Increasing the number of layers beyond two does not yield further gains and may slightly degrade performance, likely due to overfitting or increased training complexity. Conversely, a single-layer Q-Former results in a noticeable decline across all metrics, indicating limited capacity for capturing instruction-conditioned visual representations.

\clearpage
